\documentclass{article}
\usepackage{ijcai21}

\usepackage{times}
\usepackage{soul}
\usepackage{url}
\usepackage[hidelinks]{hyperref}
\usepackage[utf8]{inputenc}
\usepackage[small]{caption}
\usepackage{graphicx}
\usepackage{amsmath}
\usepackage{amsthm}
\usepackage{amsfonts}
\usepackage{booktabs}
\usepackage{algorithm}
\usepackage{algorithmic}
\usepackage{dsfont}
\usepackage{multicol}
\usepackage{multirow}
\usepackage{xcolor}

\usepackage{stfloats}

\newcommand{\figuretag}[1]{%
  \addtocounter{figure}{-1}%
  \renewcommand{\thefigure}{#1}%
}

\newcommand{\tabtag}[1]{%
\renewcommand{\thetable}{B.1}%
\addtocounter{table}{-1}}

\newtheorem{proposition}{Proposition}
\newcommand{\boldzero}{\mathbf{0}}
\newcommand{\boldone}{\mathbf{1}}
\newcommand{\laffunc}{{\rm LAF}}

\newcommand{\R}{\mathbb{R}}
\newcommand{\Rset}{\mathds{R}}

\newcommand{\N}{\mathbb{N}}

\newcommand{\boldx}{\boldsymbol x}

\title{Learning Aggregation Functions}

\author{
Giovanni Pellegrini$^{1}$\thanks{Equal Contribution. Contact Authors.}
\and
Alessandro Tibo$^{2}$\footnotemark[1]
\and
\\ Paolo Frasconi$^3$
\and
Andrea Passerini$^{1,2}$ 
\And
Manfred Jaeger$^2$
\affiliations
$^1$DISI, University of Trento \\$^2$Computer Science Department, Aalborg University\\
$^3$DINFO, Università di Firenze\\
\emails
\{giovanni.pellegrini, andrea.passerini\}@unitn.it,
\{alessandro, jaeger\}@cs.aau.dk,
paolo.frasconi@pm.me
}

\begin{document}

\maketitle

\begin{abstract}

  Learning on sets is increasingly gaining attention in the machine
  learning community, due to its widespread applicability. Typically,
  representations over sets are computed by using fixed aggregation
  functions such as sum or maximum. However, recent results showed
  that universal function representation by sum- (or max-)
  decomposition requires either highly discontinuous (and thus poorly
  learnable) mappings, or a latent dimension equal to the maximum
  number of elements in the set. To mitigate this problem, we
  introduce a learnable aggregation function (LAF) 
  for sets of arbitrary cardinality. LAF can approximate
  several extensively used aggregators (such as average, sum, maximum)
  as well as more complex functions (e.g., variance and skewness). We
  report experiments on semi-synthetic and real data showing that LAF
  outperforms state-of-the-art sum- (max-) decomposition architectures
  such as DeepSets and library-based architectures like Principal Neighborhood Aggregation, and can be effectively combined with attention-based architectures.

\end{abstract}

\section{Introduction}

The need to aggregate representations is ubiquitous in deep
learning. Some recent examples include max-over-time pooling used 
in convolutional networks for sequence
classification~\cite{DBLP:conf/emnlp/Kim14}, average pooling of
neighbors in graph convolutional networks
\cite{DBLP:conf/iclr/KipfW17}, max-pooling in Deep
Sets~\cite{deepsets2017}, in (generalized) multi-instance
learning~\cite{DBLP:journals/jmlr/TiboJF20} and in
GraphSAGE~\cite{DBLP:conf/nips/HamiltonYL17}. In all the above cases
(with the exception of LSTM-pooling in GraphSAGE) the aggregation
function is predefined, i.e., not tunable, which may be in general a
disadvantage~\cite{DBLP:conf/icml/IlseTW18}.  Sum-based aggregation
has been advocated based on theoretical findings showing the
permutation invariant functions can be
sum-decomposed~\cite{deepsets2017,xu2018how}. However, recent
results~\cite{wagstaff2019limitations} showed that this universal
function representation guarantee requires either highly discontinuous
(and thus poorly learnable) mappings, or a latent dimension equal to
the maximum number of elements in the set. This suggests that learning set
functions that are accurate on sets of large cardinality is difficult.

Inspired by previous work on learning uninorms~\cite{MelnHull16}, we
propose a new parametric family of aggregation functions that we call
LAF, for {\em learnable aggregation functions}. A single LAF unit can approximate
standard aggregators like sum, max or mean as well as model
intermediate behaviours (possibly different in different areas of the
space). In addition, LAF layers with multiple aggregation units can
approximate higher order moments of distributions like variance,
skewness or kurtosis. In contrast, other authors~\cite{corso2020principal}
suggest to employ a predefined library of elementary aggregators to be combined.
Since LAF can represent sums, it can
be seen as a smooth version of the class of functions that
are shown in~\cite{deepsets2017} to enjoy universality results in representing
set functions. The hope is that being smoother, LAF is more easily learnable.
Our empirical findings show that this can be actually the case, especially
when asking the model to generalize over large sets.
In particular, we find that:
\begin{itemize}
\item LAF layers can learn a wide range of aggregators 
(including higher-order moments) on sets of scalars without background knowledge on the
nature of the aggregation task.
\item LAF layers on the top of traditional layers can learn the
same wide range of aggregators on sets of high dimensional vectors (MNIST images).
\item LAF outperforms state-of-the-art set learning methods such as DeepSets 
and PNA on real-world problems involving point clouds and text concept set retrieval.
\item LAF performs comparably to PNA on random graph generation tasks, 
outperforming several graph neural networks architectures including     
GAT~\cite{gat} and GIN~\cite{xu2018how}.
\end{itemize}

\section{The LAF Framework}
\label{sec:laf}
\begin{table*}[t]
    \centering
      \begin{small}
        \begin{sc}
          \begin{tabular}{lc|cc|cc|cc|cc|cccc|l}
            \toprule
      Name & Definition  & $a$ &$ b$ & $c$ &$ d$ & $e$ & $f$ & $g$ &$ h$ & $\alpha$ & $\beta$ & $\gamma$ & $\delta$ & limits \\ 
            \midrule
      constant & $\kappa\in\R$ & 0  & 1 & - & - & 0 & 1 & - & - & $\kappa$ & 0 & 1 & 0  &  \\
      max & $\max_i x_i$ & $1/r$ &  $r$ & - & - & 0 & 1 & - & - & 1 & 0 & 1 & 0 & $r\rightarrow\infty$\\
      min& $\min_i x_i$  & 0 & 1 & $1/r$ &  $r$  & 0 & 1 & - & - & 1 &-1 & 1 & 0 & $r\rightarrow\infty$\\
      sum & $\sum_i x_i$ & 1 & 1 & - & - & 0 & 1 & - & - & 1 & 0 & 1 & 0 & \\
      nonzero count& $|\{i: x_i\neq 0\}|$  & 1 & 0 & - & - &  0 & 1 & - & - & 1 & 0 & 1 & 0 & \\
      mean & $1/N \sum_i x_i$ & 1 & 1  & - & - & 1 & 0 & - & - & 1 & 0 & 1 & 0 & \\
      $k$th moment & $1/N \sum_i x_i^k$ & 1 & $k$ & - & - & 1 & 0 & - & - & 1 & 0 & 1 & 0 & \\
      $l$th power of $k$th moment& $(1/N \sum_i x_i^k)^l$  & $l$ & $k$ & - & - & $l$ & 0 & - & - & 1 & 0 & 1 & 0 & \\
      min/max & $\min_i x_i/\max_i x_i $  & 0 & 1 & $1/r$ &  $r$ & $1/s$ &  $s$ & - & - & 1 & 1 & 1 & 0 & $r,s\rightarrow\infty$\\
      max/min  & $\max_i x_i/\min_i x_i $ & $1/r$ &  $r$ & - & - & 0 & 1 &  $1/s$ &  $s$ & 1 & 0 & 1 & 1&  $r,s\rightarrow\infty$\\
            \bottomrule
          \end{tabular}
        \end{sc}
      \end{small}
    \caption{Different functions achievable by varying the parameters in the formulation in Equation~\ref{eq:foursigma}.}
    \label{tab:foursigma_variants} 
  \end{table*}
We use $\boldx=\{x_1,\ldots,x_N\}$ to denote finite multisets of real numbers $x_i\in \R$. Note that directly taking
$\boldx$ to be a multiset, not  a vector, means that there is no need to  define properties like exchangeability
or permutation equivariance for operations on $\boldx$. An aggregation function $\emph{agg}$ is any
function that returns for any multiset $\boldx$ of arbitrary  cardinality $N\in\N$ a value
$\emph{agg}(\boldx)\in\R$. 

Standard aggregation functions like \emph{mean} and \emph{max} can be understood as (normalized) $L_p$-norms.
We therefore build our parametric LAF aggregator around functions of a form that generalizes $L_p$-norms:
\begin{equation}
  \label{eq:genLp}
  L_{a,b}(\boldx):=\left( \sum_i x_i^b\right)^a\hspace{5mm} (a,b \geq 0).
\end{equation}
$L_{a,b}$ is invariant under the addition of zeros: $L_{a,b}(\boldx) = L_{a,b}(\boldx\cup \boldzero)$ where $\boldzero$ is a
multiset of zeros of arbitrary cardinality.
In order to also enable aggregations that can represent \emph{conjunctive} behaviour
such as \emph{min}, we make symmetric use of aggregators of the multisets $\boldone-\boldx := \{1-x_i|x_i\in\boldx\}$.
For $L_{a,b}(\boldone-\boldx)$ to be a well-behaved, dual version of $ L_{a,b}(\boldx)$,
the values in $\boldx$ need to lie in the range $[0,1]$.
We therefore restrict the following definition of our \emph{learnable aggregation function} to sets $\boldx$
whose elements are in $[0,1]$:
\begin{equation}
  \label{eq:foursigma} 
  \laffunc(\boldx):= \frac{\alpha L_{a,b}(\boldx) + \beta L_{c,d}(\boldone-\boldx)}
  {\gamma L_{e,f}(\boldx) +\delta L_{g,h}(\boldone-\boldx) }
\end{equation}
defined by tunable parameters $a,\ldots,h\geq 0$, and $\alpha,\ldots,\delta \in\R$. In cases where sets need to
be aggregated whose elements are not already bounded by $0,1$, we apply a sigmoid function to the set elements
prior to aggregation. 

Table~\ref{tab:foursigma_variants} shows how a number of important aggregation functions are special cases of
\laffunc{} (for values in $[0,1]$). 
We make repeated use of the fact that $L_{0,1}$ returns the constant 1. 
For max and min \laffunc{} only provides an asymptotic approximation in the limit of specific
function parameters (as indicated in the last column of Table~\ref{tab:foursigma_variants}). 
 In most cases,  the parameterization of \laffunc{}
 for the functions in Table~\ref{tab:foursigma_variants} will not be unique.
 Being able to encode the powers of moments implies that e.g. the
variance of $\boldx$ can be expressed as the difference $1/N \sum_i x_i^2 - (1/N \sum_i x_i)^2$ of two
\laffunc{} aggregators. 

Since \laffunc{} includes sum-aggregation, we can adapt the results of~\cite{deepsets2017} and~\cite{wagstaff2019limitations}
on the theoretical
universality of sum-aggregation as follows.

\begin{proposition}
\label{prop:universal}
  Let ${\cal X}\subset \Rset$ be countable, and $f$ a function defined on finite multisets with
  elements from ${\cal X}$. Then there exist functions $\phi: {\cal X}\rightarrow [0,1]$,
  $\rho: \Rset \rightarrow \Rset$, and a parameterization of \laffunc{}, such that
  $f(\boldx)= \rho (LAF(\phi\boldx);\alpha,\beta,\gamma,\delta,a,b,c,d)$, where
  $\phi\boldx$ is the multiset $\{\phi(x)|x\in\boldx\}$. 
\end{proposition}

A proof in~\cite{wagstaff2019limitations} for a very similar proposition used a mapping from
${\cal X}$ into the reals. Our requirement that \laffunc{} inputs must be in $[0,1]$ requires a
modification of the  proof (contained in the supplementary material\footnote{See \url{https://github.com/alessandro-t/laf} for supplementary material and code.}), which for the definition of $\phi$ relies on
a randomized construction. 
Proposition~\ref{prop:universal} shows that we retain the
theoretical universality guarantees of~\cite{deepsets2017}, while enabling a wider
range of solutions based on
continuous encoding and decoding functions.

\begin{figure*}[ht!]
  \begin{tabular}{cc}
      \begin{minipage}{.34\linewidth}
        \includegraphics[width=1\textwidth]{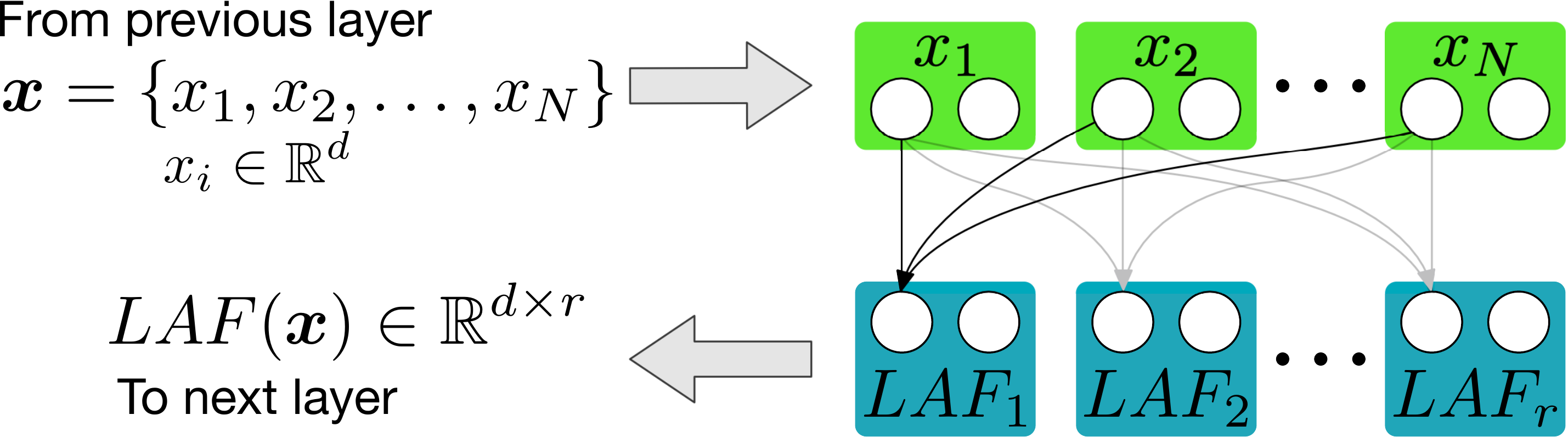}
      \end{minipage}
    &
      \begin{minipage}{.61\linewidth}
        \includegraphics[scale=0.135]{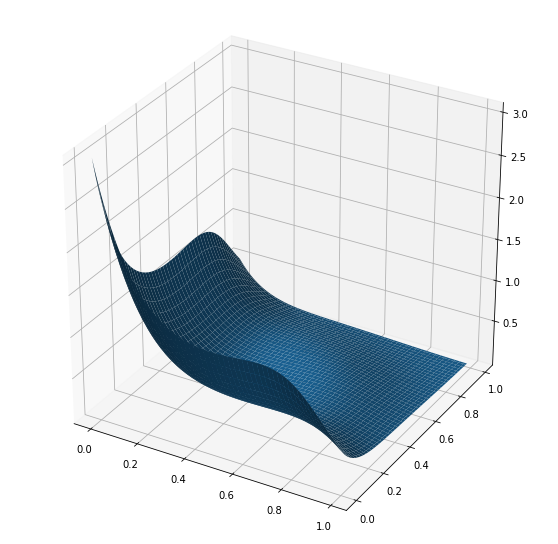}
        \includegraphics[scale=0.135]{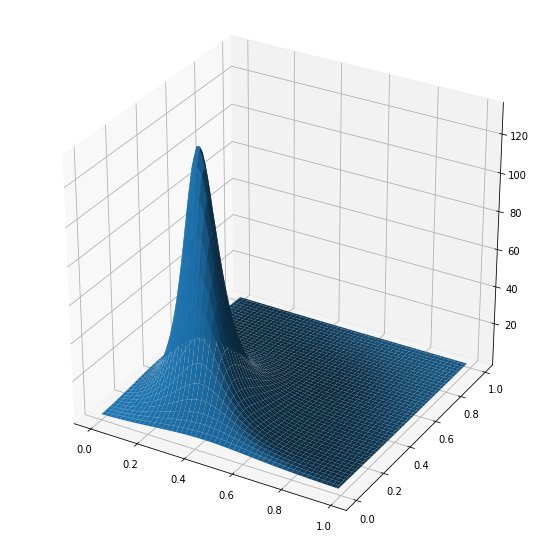}
        \includegraphics[scale=0.135]{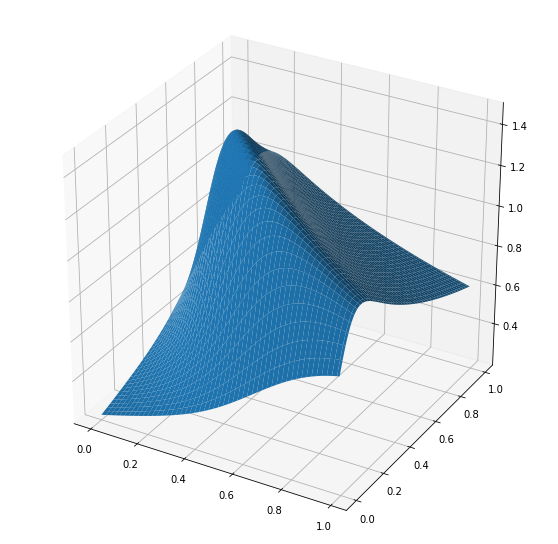}
        \includegraphics[scale=0.135]{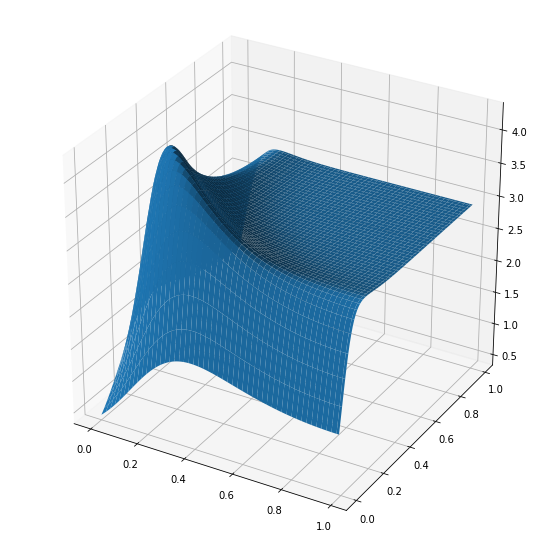}
      \end{minipage}
  \end{tabular}
\caption{Left: End-to-end LAF architecture. Right: \laffunc{} functions with randomly generated parameters.}
\label{fig:randlafs}
\end{figure*}

LAF enables a continuum of intermediate and hybrid aggregators. In Figure~\ref{fig:randlafs} we plot 4 different randomly generated 
\laffunc{} functions over $[0,1]\times [0,1]$, i.e., evaluated over 
sets of size 2. Parameters $\alpha,\ldots,\gamma$ were randomly sampled in  
$[0,1]$, $b,d,f,h$ in $\{0,\ldots,5\}$, and
$a,c,e,g$ obtained as $1/i$ with $i$ a random integer from $\{0,\ldots,5\}$.
The figure illustrates the rich repertoire of aggregation functions with different 
qualitative behaviours already for non-extreme parameter values.

Learning the functions depicted in Table~\ref{tab:foursigma_variants} can in principle be done by a single \laffunc{} unit. However, learning complex aggregation functions might require a larger number of independent units, in that the final aggregation is the result of the combination of simpler aggregations. Moreover, a \laffunc{} layer should be able to approximate the behaviour of simpler functions also when multiple units are used. Therefore, we analyzed the application of multiple \laffunc{} units to some of the known functions in Table~\ref{tab:foursigma_variants}. The details and the visual representation of this analysis is shown in the supplementary material. Although using only one function is sometimes sufficient to greatly approximate the target function, the error variance among different runs is quite high, indicating that the optimization sometimes fails to converge to a good set of parameters. Multiple units provide more stability while performing better than a single unit aggregation in some cases. We therefore construct the \laffunc{} architecture for the experimental section by using multiple aggregators, computing the final aggregation as a linear combination of the units aggregations.
Several \laffunc{} units can be combined as shown in Figure~\ref{fig:randlafs}, to capture different aspects of the input set, which can be in general a multiset of vectors $\boldx=\{x_1, \dots, x_N\}$, where now $x_i\in\mathbb{R}^d$. Note that multiple aggregators are also used in related frameworks such as DeepSets~\cite{deepsets2017} or Graph Neural Networks~\cite{gat,corso2020principal}. 
A module with $r$ \laffunc{} units takes as input $d$-dimensional vectors and produces a vector of
size $r \times d$ as output. Each LAF unit performs an \textit{element-wise} aggregation of the
vectors in the set such that $L_{k,j}=\mathrm{LAF}(\{x_{i,j},\ldots,x_{N,j}\};\alpha_k,\beta_k,\gamma_k,\delta_k,a_k,b_k,c_k,d_k)$
for $k=1,\ldots,r$ and $j=1,\ldots,d$. The output vector can be then fed into the next layer.

\section{Related Work}
\label{sec:rel}

Several studies address the problem of aggregating data over sets.
Sum-decomposition strategies have been used in~\cite{deepsets2017} for
points cloud classification and set expansion tasks and in
\cite{santoro2017simple} for question answering and dynamic physical
systems computation. Max, sum and average are standard aggregation
functions for node neighborhoods in graph neural
networks~\cite{DBLP:conf/nips/HamiltonYL17,DBLP:conf/iclr/KipfW17,xu2018how,gat}.
\cite{deepsets2017} first proved universal representation results for
these standard aggregators when combined with learned mappings over
inputs and results of the aggregation. However,
\cite{wagstaff2019limitations} showed that these universality results
are of little practical use, as they either require highly
discontinuous mappings that would be extremely difficult to learn, or
a latent dimension that is at least the size of the maximum number of
input elements. 

\textit{Uninorms}~\cite{yager1996uninorm} are a class of aggregation
functions in fuzzy logic that can behave in a \textit{conjunctive},
\textit{disjunctive} or \textit{averaging} manner depending on a
parameter called \textit{neutral element}. \cite{MelnHull16} proposed
to learn fuzzy aggregators by adjusting these learnable parameters,
showing promising results on combining reviewers scores on papers into
an overall decision of acceptance or reject. Despite the advantage of
incorporating different behaviours in one single function, uninorms
present discontinuities in the regions between aggregators, making
them not amenable to be utilized in fully differentiable
frameworks. Furthermore the range of possible behaviours is restricted
to those commonly used in the context of fuzzy-logic. 

The need for considering multiple candidate aggregators is advocated
in a very recent work that was developed in parallel with our
framework~\cite{corso2020principal}. The resulting architecture,
termed \textit{Principal Neighborhood Aggregation} (PNA) combines
multiple standard aggregators, including most of the ones we consider
in the LAF framework, adjusting their outputs with degree
scalers. However, the underlying philosophy is rather different. PNA
aims at learning to select the appropriate aggregator(s) from a pool
of candidates, while LAF explores a continuous space of aggregators
that includes standard ones as extreme cases. Our experimental
evaluation shows that PNA has troubles in learning aggregators that
generalize over set sizes, despite having them in the pool of
candidates, likely because of the quasi-combinatorial structure of its
search space. On the other hand, LAF can successfully learn even the
higher moment aggregators and consistently outperforms PNA.

Closely connected, but somewhat complementary to aggregation operators
are \emph{attention mechanisms}~\cite{DBLP:journals/corr/BahdanauCB14,%
DBLP:conf/nips/VaswaniSPUJGKP17}. They have been explored to manipulate
set data in~\cite{DBLP:conf/icml/LeeLKKCT19} and in the 
context of multi-instance learning~\cite{DBLP:conf/icml/IlseTW18}.
Attention operates at the level of set elements, and aims at a transformation
(weighting) of their representations such as to optimize a  subsequent weighted
sum-aggregation. Although the objectives of attention-based frameworks and \laffunc{} are different in principle, our method can be used inside attention frameworks as the aggregation layer of the learned representation. We discuss the combination of \laffunc{} and attention in Section~\ref{sec:stlaf} showing the advantage of using \laffunc{}.

\section{Experiments}
\label{sec:exps}
In this section, we present and discuss experimental results showing the potential of the \laffunc{} framework on both synthetic and real-world tasks. Synthetic experiments are aimed at showing the ability of \laffunc{} to learn a wide range of aggregators and its ability to generalize
over set sizes (i.e., having test-set sets whose cardinality exceeds the cardinality of the
training-set sets), something that alternative architectures based on predefined 
aggregators fail to achieve. We use DeepSets, PNA, and LSTM as representatives of these architectures. The LSTM architecture corresponds to a version of DeepSets where the aggregation function is replaced by a LSTM layer. 
Experiments on diverse tasks including point cloud classification, text concept set retrieval and graph properties prediction are aimed at showing the potential of the framework on real-world applications.

\subsection{Scalars Aggregation}\label{sec:exps:scalars}

\begin{figure*}[ht!]
\includegraphics[width=\textwidth]{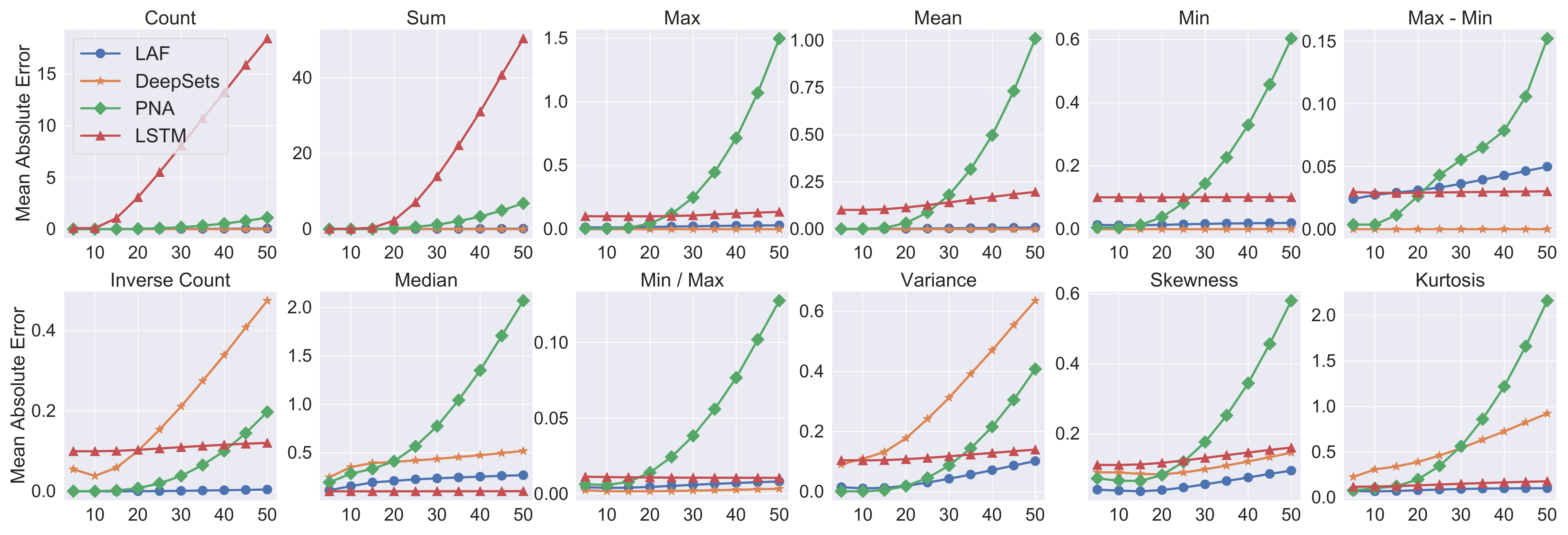}
\caption{Test performances for the synthetic experiment with integer scalars on increasing test set size. The x axis represents the maximum test set cardinality, the y axis depicts the MAE. The dot, star, diamond and triangle symbols denote LAF, DeepSets, PNA, and LSTM respectively. Skewness: $1/N \sum_i ((x_i - \hat{\mu})/\hat{\sigma})^3$, Kurtosis: $1/N \sum_i ((x_i - \hat{\mu})/\hat{\sigma})^4$, where $\hat{\mu}$ and $\hat{\sigma}$ are the sample mean and standard deviation.} 
\label{fig:int_all_small}
\end{figure*}
This section shows the learning capacity of the \laffunc{} framework to learn simple
and complex aggregation functions where constituents of the sets are simple
numerical values. 
In this setting we consider sets made of scalar integer values.
The training set is constructed as follows: for each set, we initially sample its cardinality $K$ from a uniform distribution taking values in $\{2, \dots, M\}$, and then we uniformly sample $K$ integers in $\{0,\ldots,9\}$. For the training set we use $M=10$. We construct several test sets for different values of $M$ ($M=5,10,15,20,25,30,35,40,45,50$). This implies that models need to generalize to larger set sizes. Contrarily to the training set, each test set is constructed in order to diversify the target labels it contains, so as to avoid degenerate behaviours for large set sizes (e.g., maximum constantly equal to 9). Each synthetic dataset is composed of 100,000 sets for training,
20,000 set for validating and 100,000 for testing. 
The number of aggregation units is set as follows. The model contains nine \laffunc{} (Equation~\ref{eq:foursigma}) units, whose parameters $\{a_k,\dots,h_k\}$, $k=1,\ldots,9$ are initialized according to a uniform sampling in $[0,1]$ as those parameters must be positive, whereas the coefficients $\{\alpha, \dots, \delta \}$ are initialized with a Gaussian distribution with zero mean and standard deviation of 0.01 to cover also negative values.
The positivity constraint for parameters $\{a,b,\ldots,h\}$ is enforced by projection during the optimization process. The remaining parameters can take on negative values. 
DeepSets also uses nine units: three max units, three sum units, and three mean
units and PNA uses seven units: mean, max, sum, standard deviation, variance,
skewness and kurtosis. Preliminary experiments showed that expanding the set of
aggregators for PNA with higher order moments only leads to worse performance.
Each set of integers is fed into an embedding layer (followed by a sigmoid)
before performing the aggregation function. DeepSets and PNA do need an
embedding layer (otherwise they would have no parameters to be tuned).
Although \laffunc{} does not need an embedding layer, we used it in all models
to make the comparison more uniform.
The architecture details are reported in the supplementary material.
We use the Mean Absolute Error (MAE) as a loss function to calculate the prediction error.

Figure \ref{fig:int_all_small} shows the trend of the MAE for the three methods for increasing test set sizes, for different types of target aggregators. As expected, DeepSets manages to learn the identity function and thus correctly models aggregators like sum, max and mean. Even if \laffunc{} needs to adjust its parameters in order to properly aggregate the data, its performance are competitive with those of DeepSets. When moving to more complex aggregators like inverse count, median or moments of different orders, DeepSets fails to learn the latent representation. One the other hand, the performance of \laffunc{} is very stable for growing set sizes. While having in principle at its disposal most of the target aggregators (including higher order moment) PNA badly overfits over the cardinality of sets in the training set in all cases (remember that the training set contains sets of cardinality at most 10). The reason why \laffunc{} substantially outperforms PNA on large set sizes could be explained in terms of a greater flexibility to adapt to the learnt representation. Indeed, \laffunc{} parameters can adjust the \textit{laf} function to be compliant with the latent representation even if the input mapping fails to learn the identity. On the other hand, having a bunch of fixed, hard-coded aggregators, PNA needs to be able to both learn the identity mapping and select the correct aggregator among the candidates. Finally, LSTM results are generally poor when compared to the other methods, particularly in the case of the count and the sum.

\subsection{MNIST Digits}\label{sec:exps:mnist}
We performed an additional set of experiments aiming to demonstrate
the ability of \laffunc{} to learn from more complex representations of the
data by plugging it into end-to-end differentiable architectures.  In
these experiments, we thus replaced numbers by visual representations
obtained from MNIST digits. Unlike the model proposed in the previous
section, here we use three dense layers for learning picture
representations before performing the aggregation function. Results
obtained in this way are very similar to those obtained with numerical
inputs and due to space limitations we report them along with other
architectural and experimental details in the supplementary material.

\subsection{Point Clouds}\label{sec:exps:point:cloud}
In order to evaluate \laffunc{} on real-world dataset, we consider point cloud classification, a prototype task for set-wise prediction. Therefore, we run experimental comparisons on the ModelNet40~\cite{wu20153d} dataset, which consists of 9,843 training and 2,468 test point clouds of objects distributed over 40 classes. The dataset is preprocessed following the same procedure described by~\cite{deepsets2017}. 
We create point clouds of 100 and 1,000 three-dimensional points by adopting the point-cloud library's sampling routine developed by~\cite{rusu20113d} and normalizing each set of points to have zero mean (along each axis) and unit (global) variance. We refer with P100 and P1000 to the two datasets. For all the settings, we consider the same architecture and hyper-parameters of the DeepSets permutation invariant model described by~\cite{deepsets2017}. For \laffunc{}, we replace the original aggregation function (max) used in DeepSets with 10 \laffunc{} units, while for PNA we use the concatenation of max, min, mean, and standard deviation, as proposed by the authors. For PNA we do not consider any scaler, as the cardinalities of the sets are fixed. 
\begin{table}[t]
    \centering
      \begin{small}
        \begin{sc}
          \begin{tabular}{lcc}
            \toprule
            Method & P100  & P1000 \\
            \midrule
            DeepSets & 82.0$\pm$2.0\% & \textbf{87.0$\pm$1.0}\% \\
            PNA      & 82.9$\pm$0.7\% & 86.4$\pm$0.6\% \\
            LSTM     & 78.7$\pm$1.1\% & 82.2$\pm$1.7\% \\
            LAF      & \textbf{84.0$\pm$0.6}\% & \textbf{87.0$\pm$0.5}\% \\
            \bottomrule
          \end{tabular}
        \end{sc}
      \end{small}
    \caption{Results on the Point Clouds classification task. Accuracies with standard deviations (over 5 runs) for the ModelNet40 dataset.}
    \label{tab:point:cloud}
  \end{table}
 \begin{table*}[bp!]
    \centering
      \begin{small}
        \begin{sc}
        \setlength{\tabcolsep}{3.5pt}
          \begin{tabular}{l|ccccc|ccccc|ccccc}
            \toprule
            \multirow{3}[1]{*}{Method} & \multicolumn{5}{c|}{LDA-1$k$ (Vocab = $17k$)}
            & \multicolumn{5}{c|}{LDA-3$k$ (Vocab = $38k$)} & \multicolumn{5}{c}{LDA-5$k$ (Vocab = $61k$)} \\

            & \multicolumn{3}{c}{\textbf{Recall(\%)}} & \multirow{2}[1]{*}{\textbf{MRR}} & \multirow{2}[1]{*}{\textbf{Med.}} 
            & \multicolumn{3}{c}{\textbf{Recall(\%)}} & \multirow{2}[1]{*}{\textbf{MRR}} & \multirow{2}[1]{*}{\textbf{Med.}}
            & \multicolumn{3}{c}{\textbf{Recall(\%)}} & \multirow{2}[1]{*}{\textbf{MRR}} & \multirow{2}[1]{*}{\textbf{Med.}}\\
            & \textbf{@10} & \textbf{@100} & \textbf{@1k} & &
            & \textbf{@10} & \textbf{@100} & \textbf{@1k} & &
            & \textbf{@10} & \textbf{@100} & \textbf{@1k} & & \\
            \midrule
            Random & 0.06 & 0.6 & 5.9 & 0.001 & 8520 & 0.02 & 0.2 & 2.6 & 0.000 & 28635 & 0.01 & 0.2 & 1.6 & 0.000 & 30600 \\
            Bayes Set & 1.69 & 11.9 & 37.2 & 0.007 & 2848 & 2.01 & 14.5 & 36.5 & 0.008 & 3234 & 1.75 & 12.5 & 34.5 & 0.007 & 3590 \\
            w2v Near & 6.00 & \textbf{28.1} & 54.7 & 0.021 & 641 & 4.80 & 21.2 & 43.2 & 0.016 & 2054 & 4.03 & 16.7 & 35.2 & 0.013 & 6900 \\
            NN-max & 4.78 & 22.5 & 53.1 & 0.023 & 779 & 5.30 & 24.9 & 54.8 & 0.025 & 672 & 4.72 & 21.4 & 47.0 & 0.022 & 1320 \\ 
            NN-sum-con & 4.58 & 19.8 & 48.5 & 0.021 & 1110 & 5.81 & 27.2 & 60.0 & 0.027 & 453 & 4.87 & 23.5 & 53.9 & 0.022 & 731 \\
            NN-max-con & 3.36 & 16.9 & 46.6 & 0.018 & 1250 & 5.61 & 25.7 & 57.5 & 0.026 & 570 & 4.72 & 22.0 & 51.8 & 0.022 & 877 \\ 
            DeepSets & 5.53 & 24.2 & 54.3 & 0.025 & 696 & 6.04 & 28.5 & 60.7 & 0.027 & 426 & 5.54 & 26.1 & 55.5 & 0.026 & 616 \\
            \midrule
            DeepSets$^*$ & 5.89  & 26.0 & \textbf{55.3} & 0.026 & \textbf{619} & 7.56 & 28.5 & \textbf{64.0} & 0.035 & 349 & 6.49 & 27.9 & \textbf{56.9} & 0.030 & 536 \\
            PNA & 5.56 & 24.7 & 53.2 & 0.027 & 753 & 7.04 & 27.2 & 58.7 & 0.028 & 502 & 5.47 & 23.8 & 52.4 &  0.025 & 807 \\
            LSTM & 4.29 & 21.5 & 52.6 & 0.022 & 690 & 5.56 & 25.7 & 58.8 & 0.026 & 830 & 4.87 & 23.8 & 55.0 & 0.022 & 672 \\
            LAF & \textbf{6.51} & 26.6 & 54.5 & \textbf{0.030} & 650 & \textbf{8.14} & \textbf{32.3} & 62.8 & \textbf{0.037} & \textbf{339} & \textbf{6.71} & \textbf{28.3} & \textbf{56.9} & \textbf{0.031} & \textbf{523} \\
            \bottomrule
          \end{tabular}
        \end{sc}
      \end{small}
    \caption{Results on Text Concept Set Retrieval on LDA-1k, LDA-3k, and LDA-5k. Bold values denote the best performance for each metric.}
    \label{tab:set:expansion}
  \end{table*}
Results in Table~\ref{tab:point:cloud} show that \laffunc{} produces an advantage in the lower resolution dataset (i.e., on P100), while it obtains comparable (and slightly more stable) performances in the higher resolution one (i.e., on P1000). These results suggest that having predefined aggregators is not necessarily an optimal choice in real world cases, and that the 
flexibility of \laffunc{} in modeling diverse aggregation functions can boost performance and stability.

\subsection{Set Expansion}
Following the experimental setup of DeepSets, we also considered the \textit{Set Expansion}
task. In this task the aim is to augment a set of objects of the same class with other similar objects, as explained in~\cite{deepsets2017}. The model learns to predict a score for an object given a query set and decide whether to add the object to the existing set. Specifically, \cite{deepsets2017} consider the specific application of set expansion to text concept retrieval. The idea is to retrieve words that belong to a particular concept, giving as input set a set of words having the same concept. We employ the same model and hyper-parameters of the original publication, where we replace the sum-decomposition aggregation with LAF units for our methods and the min, max, mean, and standard deviation aggregators for PNA.

We trained our model on sets constructed from a vocabulary of different size, namely \textit{LDA-1K}, \textit{LDA-3K} and \textit{LDA-5K}. Table~\ref{tab:set:expansion} shows the results of \laffunc{}, DeepSets and PNA on different evaluation metrics. We report the retrieval metrics recall@K, median rank and mean reciprocal rank. We also report the results on other methods the authors compared to in the original paper. More details on the other methods in the table can be found in the original publication.  Briefly, \textit{Random} samples a word uniformly from the vocabulary; \textit{Bayes Set}~\cite{ghahramani2006bayesian}; \textit{w2v-Near} computes the nearest neighbors in the
word2vec~\cite{mikolov2013distributed} space;  \textit{NN-max} uses a similar architecture as our DeepSets but uses max pooling to compute the set feature, as opposed to sum pooling; \textit{NN-max-con} uses max pooling on set elements but concatenates this pooled representation with that of query for a final set feature; \textit{NN-sum-con} is similar to NN-max-con but uses sum pooling followed by concatenation with query representation. For the sake of fairness, we have rerun DeepSets using the current implementation from the authors (indicated as DeepSet$^{*}$ in Table~\ref{tab:set:expansion}), exhibiting better results than the ones reported in the original paper. Nonetheless, \laffunc{} outperforms all other methods in most cases, especially on \textit{LDA-3K} and \textit{LDA-5K}.

\subsection{Multi-task Graph Properties}
\cite{corso2020principal} defines a benchmark consisting of 6 classical graph theory tasks on artificially generated graphs from a wide range of popular graph types like Erdos-Renyi, Barabasi-Albert or star-shaped graphs. Three of the tasks are defined for nodes, while the other three for whole graphs. The node tasks are the single-source shortest-path lengths (N1), the eccentricity (N2) and the Laplacian features (N3). The graph tasks are graph connectivity (G1), diameter (G2), and the spectral radius (G3).
We compare \laffunc{} against PNA by simply replacing the original PNA aggregators and scalers with 100 \laffunc{} units (see Equation $\ref{eq:foursigma}$). Table~\ref{tab:pna:benchmark} shows that albeit these datasets were designed to highlight the features of the PNA architecture, that outperforms a wide range of alternative graph neural network approaches \laffunc{} produces competitive results, outperforming state-of-the-art GNN approaches like GIN~\cite{xu2018how}, GCN~\cite{DBLP:conf/iclr/KipfW17} and GAT~\cite{gat} and even improving over PNA on spectral radius prediction. PNA$^*$ is the variant without scalers of PNA still proposed by \cite{corso2020principal}.
\begin{table}[t!]
    \setlength\tabcolsep{4pt}
    \centering
      \begin{small}
        \begin{sc}
          \begin{tabular}{lcccccc}
            \toprule
            Method & N1 & N2 & N3 & G1 & G2 & G3 \\
            \midrule
            Baseline & -1.87 & -1.50 & -1.60 & -0.62 & -1.30 & -1.41 \\
            GIN  & -2.00 & -1.90 & -1.60 & -1.61 & -2.17 & -2.66 \\
            GCN & -2.16 & -1.89 & -1.60 & -1.69 & -2.14 & -2.79 \\
            GAT & -2.34 & -2.09 & -1.60 & -2.44 & -2.40 & -2.70 \\
            MPNN (max) & -2.33 & -2.26 & -2.37 & -1.82 & -2.69 & -3.52 \\
            MPNN (sum) & -2.36 & -2.16 & -2.59 & -2.54 & -2.67 & -2.87 \\
            PNA$^*$ & -2.54 & -2.42 & -2.94 & \textbf{-2.61} & -2.82 & -3.29 \\
            PNA & \textbf{-2.89} & \textbf{-2.89} & \textbf{-3.77} & \textbf{-2.61} & \textbf{-3.04} & -3.57 \\
            \midrule
            LAF & -2.13 & -2.20 & -1.67 & -2.35 & -2.77 & \textbf{-3.63} \\
            \bottomrule
          \end{tabular}
        \end{sc}
      \end{small}
    \caption{Results on the Multi-task graph properties prediction benchmark, expressed in $\log10$ of mean squared error.}
    \label{tab:pna:benchmark}
  \end{table}

\section{SetTransformer With LAF Aggregation}
\label{sec:stlaf}
\begin{figure}[t!]
\includegraphics[width=0.47\textwidth]{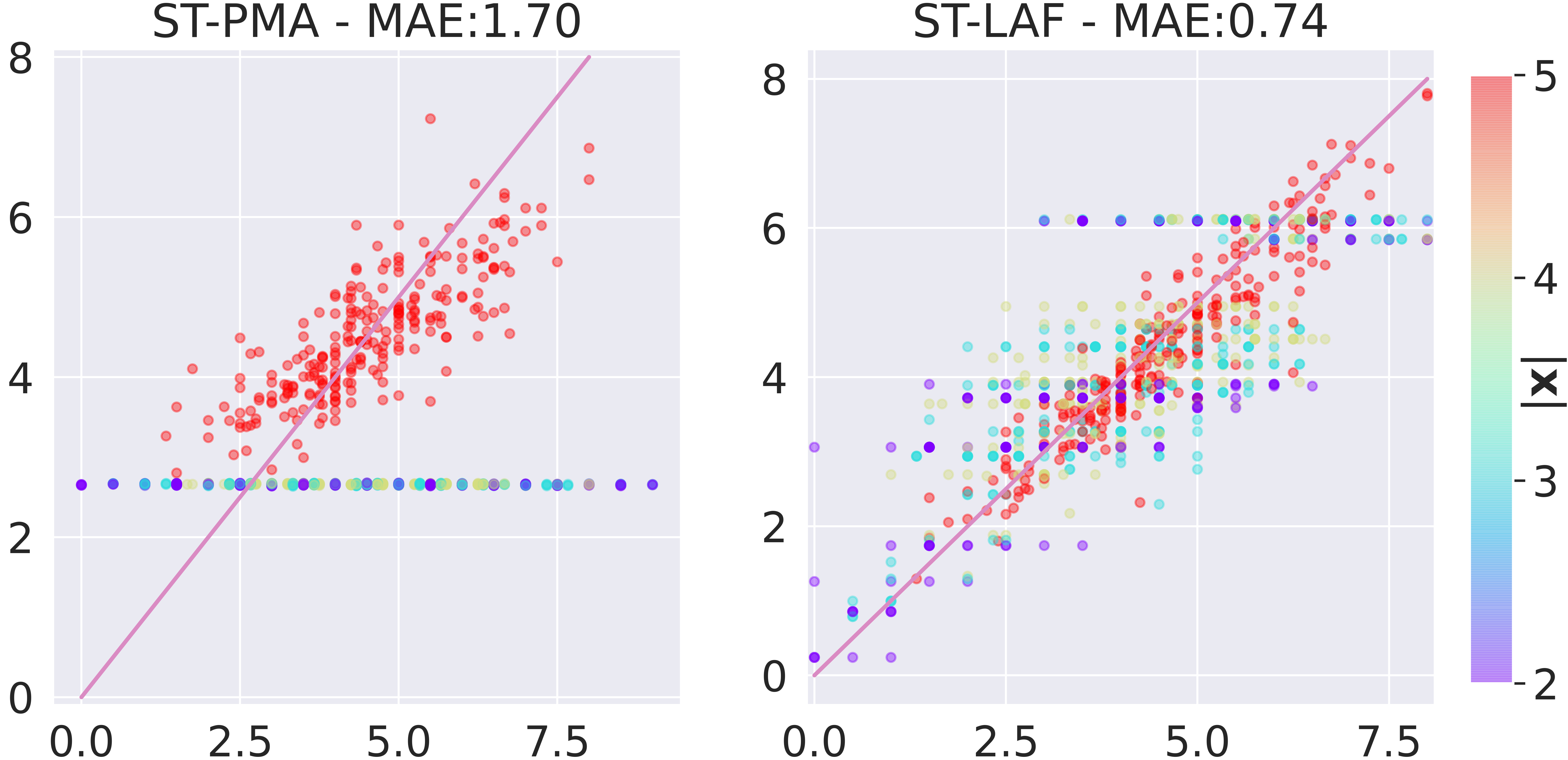}
\caption{Distribution of the predicted values for ST-PMA and ST-LAF by set cardinalities. On the x-axis the true labels of the sets, on the y-axis the predicted ones. Different colors represent the sets' cardinalities $|\boldx|$.}
\label{fig:st_pma_laf}
\end{figure}

In this section we investigate the combination of \laffunc{} aggregation with attention mechanisms on sets as proposed in the SetTransformer framework~\cite{DBLP:conf/icml/LeeLKKCT19}.
Briefly, SetTransformer consists of an \textit{encoder} and a \textit{decoder}. The encoder maps a set of input vectors into a set of feature vectors by leveraging attention blocks. The decoder employs a \textit{pooling multihead attention} (PMA) layer, which aggregates the set of feature vectors produced by the encoder. In the following experiment we replace PMA by a LAF layer. 
Inspired by one of the tasks described in \cite{DBLP:conf/icml/LeeLKKCT19}, we propose here to approximate the average of the unique numbers in a set of MNIST images. Solving the task requires to learn a cascade of two processing steps, one that detects unique elements in a set (which can be done by the transformer encoder, as shown in the experiment by~\cite{DBLP:conf/icml/LeeLKKCT19}) and one that aggregates the results by averaging (which LAF is supposed to do well). The set cardinalities are uniformly sampled from $\{2,3,4,5\}$ and each set label is calculated as the average of the unique digits contained in the set. We trained two SetTransformer models: one with PMA (ST-PMA) and the other with \laffunc{} (ST-LAF). The full implementation details are reported in the supplementary material. Quantitative and qualitative results of the evaluation are shown in Figure~\ref{fig:st_pma_laf}, where we report the MAE for both methods\footnote{We run several experiments by changing the number of seeds $k$ of PMA. All of them exhibited the same behaviour. For this experiment we used $k=1$.}. ST-\laffunc{} exhibits a nice improvement over ST-PMA for this particular task. Note that for ST-PMA only $25\%$ of the sets (red points in the scatter plot), corresponding to sets with maximum cardinality, approximates well the average, while for all other cardinalities (the remaining 75\% of the sets) ST-PMA predicts a constant value equal to the average label in the training set. ST-LAF instead clearly captures the distribution of the labels, generalizing better with respect to the set sizes. A similar behaviour was observed when learning to predict the sum rather than the average of the unique digits in a set (see supplementary material for the results).

\section{Conclusions}
The theoretical underpinnings for sum aggregation as a universal framework for defining set 
functions do not necessarily provide a template for practical solutions.  
Therefore we introduced a framework for learning aggregation functions that makes use of a
parametric aggregator to effectively explore a rich space of possible aggregations. \laffunc{} defines a new class of aggregation functions, which include as special cases widely used aggregators,
and also has the ability to learn complex functions such as higher-order moments. 
We empirically showed the generalization ability of our method on synthetic settings as well as
real-world datasets, providing comparisons with state-of-the-art sum-decomposition approaches
and recently introduced techniques. 
The flexibility of our model is a crucial aspect for potential practical use in many deep
learning architectures, due to its ability to be easily plugged into larger
architectures, as shown in our experiments with the SetTransformer.
The portability of \laffunc{} opens a new range of possible applications for aggregation functions in machine learning methods, and future research in this direction can enhance the expressivity 
of many architectures and models that deal with unstructured data.
\label{sec:conc}

\section*{Acknowledgements}
The work of PF was partly supported by the PRIN 2017 project RexLearn,
funded by the Italian Ministry of Education, University and Research, GA No 2017TWNMH2. The work of AP was partly supported by the TAILOR project, funded by EU Horizon 2020 research and innovation programme, GA No 952215. The scholarship of GP is funded by TIM and EIT Digital.

\clearpage
\bibliographystyle{named}


\clearpage
\appendix

\section{Proof of Proposition 1}
Let ${\cal X}=\{x_0,x_1,\ldots\}$. For $i\geq 0$ let $r_i$ be a random number sampled uniformly from the
interval $[0,1]$. Define $\phi(x_i):=r_i$. Let $\boldx=\{ a_i:x_i | i\in J\},  \boldx'=\{ a'_h:x_h | h\in J'\}$ be two
finite multisets with elements from ${\cal X}$, where $J,J'$ are finite index sets, and $a_i,a_h'$ denote the multiplicity
with which elements $x_i,x_h$ appear in $\boldx$, respectively $\boldx'$. Now assume that $\boldx\neq\boldx'$, but
\begin{equation}
\tag{A.1}
\label{eq:proof1} 
  \sum_{i\in J} a_i \phi(x_i) = \sum_{h\in J'} a_h' \phi(x_h), 
\end{equation}
i.e.,
\begin{equation}
\tag{A.2}
\label{eq:proof2}  
  \sum_{j\in J\cup J'} (a_j-a'_j)r_j=0,
\end{equation}
where now $a_j$, respectively $a_j'$ is defined as 0 if $j\in J'\setminus J$, respectively $j\in J\setminus J'$.
Since $\boldx\neq\boldx'$, the left side of this equation is not identical zero. Without loss of generality, we may
actually assume that all coefficients $a_j-a'_j$ are nonzero. The event that the randomly sampled values
$\{r_j| j\in J\cup J'\}$ satisfy the linear constraint (\ref{eq:proof2}) has probability zero. Since the set of
pairs of finite multisets over ${\cal X}$ is countable, also the probability that there exists any pair $\boldx\neq\boldx'$ for
which  (\ref{eq:proof1}) holds is zero. Thus, with probability one, the mapping from multisets $\boldx$ to their
sum-aggregation $\sum_{x\in\boldx}\phi(x)$ is injective. In particular, there exists a set of fixed values
$r_0,r_1,\ldots$, such that the (deterministic) mapping $x_i\mapsto r_i$ has the desired properties. The existence of the
``decoding'' function $\rho$ is now guaranteed as in the proofs of~\cite{deepsets2017,wagstaff2019limitations}.

Clearly, due to the randomized construction, the theorem and its proof have limited implications in practice. This however,
already is true for previous results along these lines, where at least for the decoding function $\rho$, not much more
than pure existence could be demonstrated.

\section{Learning}
\begin{figure*}[ht!]
\centering
\includegraphics[width=0.98\textwidth]{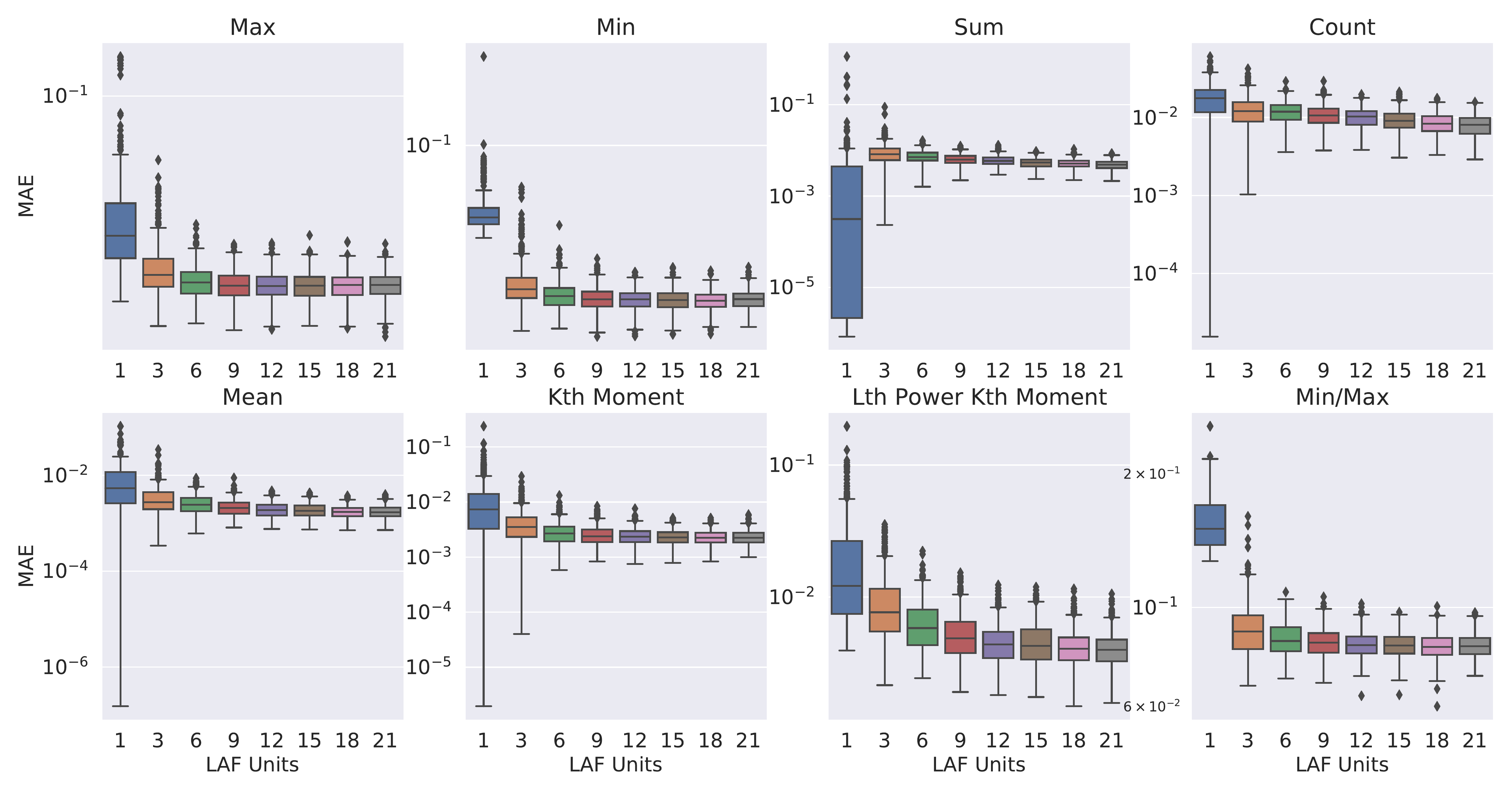}
\figuretag{B.1}
\caption{
Trend of the MAE obtained with an increasing number of LAF units for most of the functions reported in Table 1. The error distribution is obtained performing 500 runs with different random parameter initializations. A linear layer is stacked on top of the LAF layer with more than 1 unit. The y axis is plot in logaritmic scale.}
\label{fig:box_plots}
\end{figure*}

We study here the difficulty of solving the optimization problem when varying 
the number of LAF units, aiming to show that the use of multiple units helps 
finding a better solution. We formulate as learning tasks some of the target 
functions described in Table 1. Additionally, we inspect the parameters
of the learned model. 
We construct a simple architecture similar to the aggregation layer presented in Section 4, in which the aggregation is performed using one or more LAF units and, in the case of multiple aggregators, their outputs are combined together using a linear layer. 
We also discard any non-linear activation function prior to the aggregation because the input sets are composed of real numbers in the range $[0,1]$, with a maximum of 10 elements for each set. We consider 1,3,6,9,12,15,18 and 21 LAF units in this setting. For each function and for each number of units we performed 500 random restarts. The results are shown in Figure~\ref{fig:box_plots}, where we report the MAE distributions. Let's initially consider the cases when a single unit performs the aggregation. Note first that the functions listed in Table 1 can be parametrized in an infinite number of alternative ways. 
For instance, consider the \textit{sum} function. A possible solution is obtained if $L_{a,b}$ learns the \textit{sum}, $L_{e,f}=1$ and $\alpha=\gamma$. If instead $L_{a,b}=\textit{sum}$ and $L_{e,f}=L_{g,h}=1$, it is sufficient that $\gamma+\delta=\alpha$ to still obtain the sum.
This is indeed what we found when inspecting the best performing models among the various restarts, as shown in the following:
$$
sum: \frac{1.75(\sum x^{1.00})^{1.00}\ + \ 0.00(\sum (1-x)^{0.00})^{0.56}}{0.91(\sum x^{0.24})^{0.00}\ + \ 0.84(\sum (1-x)^{0.36})^{0.00}}
$$
$$
count: \frac{1.01(\sum x^{0.00})^{0.99}\ + \ 0.94(\sum (1-x)^{0.00})^{1.01}}{1.08(\sum x^{0.47})^{0.00}\ + \ 0.88(\sum (1-x)^{1.02})^{0.00}}
$$
$$
mean: \frac{1.51(\sum x^{1.00})^{1.00}\ + \ 0.00(\sum (1-x)^{0.62})^{0.00}}{0.00(\sum x^{0.30})^{0.00}\ + \ 1.51(\sum (1-x)^{0.00})^{1.00}} 
$$
A detailed overview of the parameters' values learned using one \laffunc{} unit is depicted in Table~\ref{tab:foursigma_params}. For each function in Figure~\ref{fig:box_plots}, we report the values of the random restart that obtained the lowest error.
The evaluation clearly shows that learning a function with just one LAF unit is not trivial. In some cases LAF was able to almost perfectly match the target function, but to be reasonably confident to learn a good representation many random restarts are needed, since the variance among different runs is quite large. The error variance reduces when more than one LAF unit is adopted, drastically dropping when six units are used in parallel, still maintaining a reasonable average error. Jointly learning multiple LAF units and combining their outputs can lead to two possible behaviours giving rise to an accurate approximation of the underlying function: in the first case, it is possible that one ``lucky'' unit learns a parametrization close to the target function, leaving the linear layer after the aggregation to learn to choose that unit or to rescale its output. In the second case the target function representation is ``distributed'' among the different units, here the linear layer is responsible to obtain the function by combining the LAF aggregation outputs. In the following we show another example of a learnt model, for a setting with three LAF units. Here the target function is the \textit{count}.
$$
unit1: \frac{0.81(\sum x^{0.87})^{0.37}\ + \ 0.80(\sum (1-x)^{0.74})^{0.72}}{1.19(\sum x^{0.19})^{0.72}\ + \ 1.18(\sum (1-x)^{0.00})^{0.62}}\\
$$
$$
unit2: \frac{1.43(\sum x^{0.00})^{1.10}\ + \ 1.31(\sum (1-x)^{0.01})^{0.74}}{0.64(\sum x^{0.85})^{0.00}\ + \ 0.62(\sum (1-x)^{0.46})^{0.00}}\\
$$
$$
unit3: \frac{0.83(\sum x^{0.87})^{0.37}\ + \ 0.77(\sum (1-x)^{0.12})^{0.00}}{1.17(\sum x^{0.69})^{0.86}\ + \ 1.22(\sum (1-x)^{0.00})^{0.16}}\\
$$

\begin{align*}
        &linear:& 0.02 + (-0.13*unit1)+\\
        &&+(0.50*unit2)+(-0.07*unit3)    
\end{align*}

In this case, the second unit learns a function that counts twice the elements of the set. The output of this unit is then halved by the linear layer, which gives very little weights to the outputs of the other units.
\begin{table*}[h!]
\begin{center}
\begin{small}
\begin{sc}
\begin{tabular}{l|cc|cc|cc|cc|cccc}
\toprule
Name & $a$ &$ b$ & $c$ &$ d$ & $e$ & $f$ & $g$ &$ h$ & $\alpha$ & $\beta$ & $\gamma$ & $\delta$ \\
\midrule
max & 0.28  & 4.74 & 0.00 & 0.57 & 0.33 & 1.74 & 0.00 & 0.48 & 1.68 & 0.00 & 0.90 & 0.75 \\
min & 0.28  & 0.28 & 0.27 & 1.13 & 0.30 & 0.35 & 0.87 & 3.69 & 0.51 & 0.00 & 0.45 & 1.91 \\
sum & 1.00  & 1.00 & 0.56 & 0.00 & 0.00 & 0.24 & 0.00 & 0.36 & 1.75 & 0.00 & 0.91 & 0.84 \\
count & 0.99  & 0.00 & 1.01 & 0.00 & 0.00 & 0.47 & 0.00 & 1.02 & 1.01 & 0.94 & 1.08 & 0.88 \\
mean & 1.00  & 1.00 & 0.00 & 0.62 & 0.00 & 0.30 & 1.00 & 0.00 & 1.51 & 0.00 & 0.00 & 1.51 \\
$k$th moment & 1.00  & 2.00 & 0.00 & 0.13 & 1.00 & 0.00 & 1.00 & 0.00 & 1.67 & 0.00 & 0.83 & 0.84 \\
$l$th power of $k$th moment & 2.87  & 2.15 & 0.00 & 0.91 & 2.94 & 0.00 & 1.71 & 0.00 & 1.65 & 0.01 & 1.44 & 0.24 \\
min/max & 0.06  & 0.00 & 1.52 & 2.36 & 0.18 & 4.40 & 0.64 & 7.25 & 0.23 & 0.10 & 0.27 & 2.26 \\
\bottomrule
\end{tabular}
\end{sc}
\end{small}
\end{center}
\tabtag{B.1}
\caption{Parameters' values learned with one LAF unit.}
\label{tab:foursigma_params}
\end{table*}

\section{Details of Sections~4.1 - Experiments on Scalars}

We used mini-batches of 64 sets and trained the models for 100 epochs. We use Adam as parameter optimizer, setting the initial learning rate to $1e^{-3}$ and apply adaptive decay based on the validation loss. 

Each element in the dataset is a set of scalars $\boldsymbol{x} = \{x_1,\ldots, x_N \}$, $x_i \in \mathbb{R}$. 

Network architecture:
\begin{align*}
  \boldsymbol{x}  & \rightarrow  \textsc{Embedding(10,10)} \rightarrow \textsc{Sigmoid}\\
&  \rightarrow  \textsc{LAF(9)} \rightarrow \textsc{Dense(10 $\times$ 9, 1)}
\end{align*}

\section{Details of Sections~4.2 - MNIST Digits}
In this section, we modify the experimental setting in Section~4.1 for the integers scalars to
process MNIST images of digits.  The dataset is the same as in the
experiment on scalars, but integers are replaced by randomly sampling
MNIST images for the same digits. Instances for the training and test
sets are drawn from the MNIST training and test sets, respectively.
We used mini-batches of 64 sets and trained the models for 100 epochs. We use Adam as parameter optimizer, setting the initial learning rate to $1e^{-3}$ and apply adaptive decay based on the validation loss. 
Each element in the dataset is a set of vectors $\boldsymbol{x} = \{x_1,\ldots, x_N \}$, $x_i \in \mathbb{R}^{784}$. 
Network architecture:
\begin{align*}
  \boldsymbol{x}  & \rightarrow  \textsc{Dense(784,300)} \rightarrow \textsc{Tanh}\\
& \rightarrow  \textsc{Dense(300,100)} \rightarrow \textsc{Tanh}\\  
& \rightarrow  \textsc{Dense(100,30)} \rightarrow \textsc{Sigmod}\\  
&  \rightarrow  \textsc{LAF(9)} \rightarrow \textsc{Dense(30 $\times$ 9, 1000)}\rightarrow \textsc{Tanh}\\  
& \rightarrow  \textsc{Dense(1000,100)} \rightarrow \textsc{Tanh} \rightarrow  \textsc{Dense(100,1)}
\end{align*}

\begin{figure*}
\includegraphics[width=0.98\textwidth]{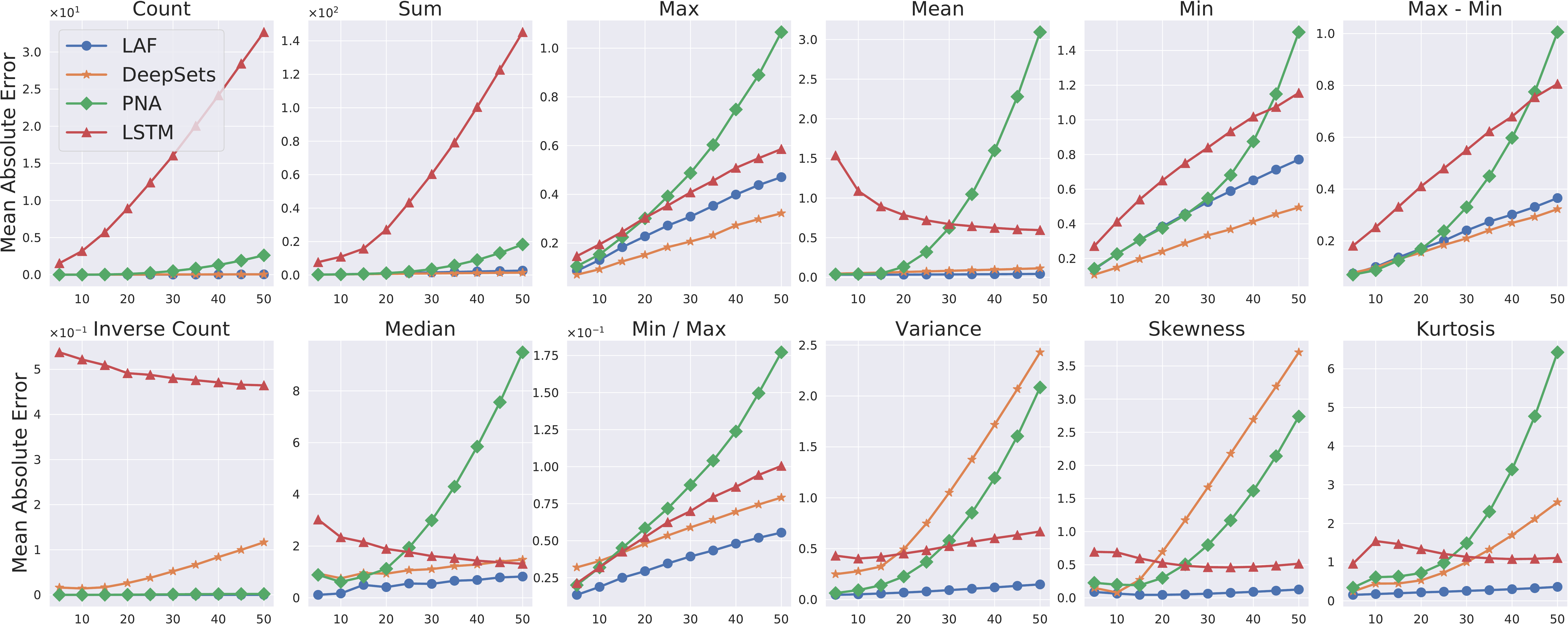}
\figuretag{D.1}
\caption{Test performances for the synthetic experiment on MNIST digits on increasing test set size. The x axis of the figures represents the maximum test set cardinality, whereas the y axis depicts the MAE. The dot, star, diamond and traingle symbols denote LAF, DeepSets, PNA and LSTM respectively.}
\label{fig:int_all_big}
\end{figure*}
\begin{figure*}[ht!]
\includegraphics[width=0.98\textwidth]{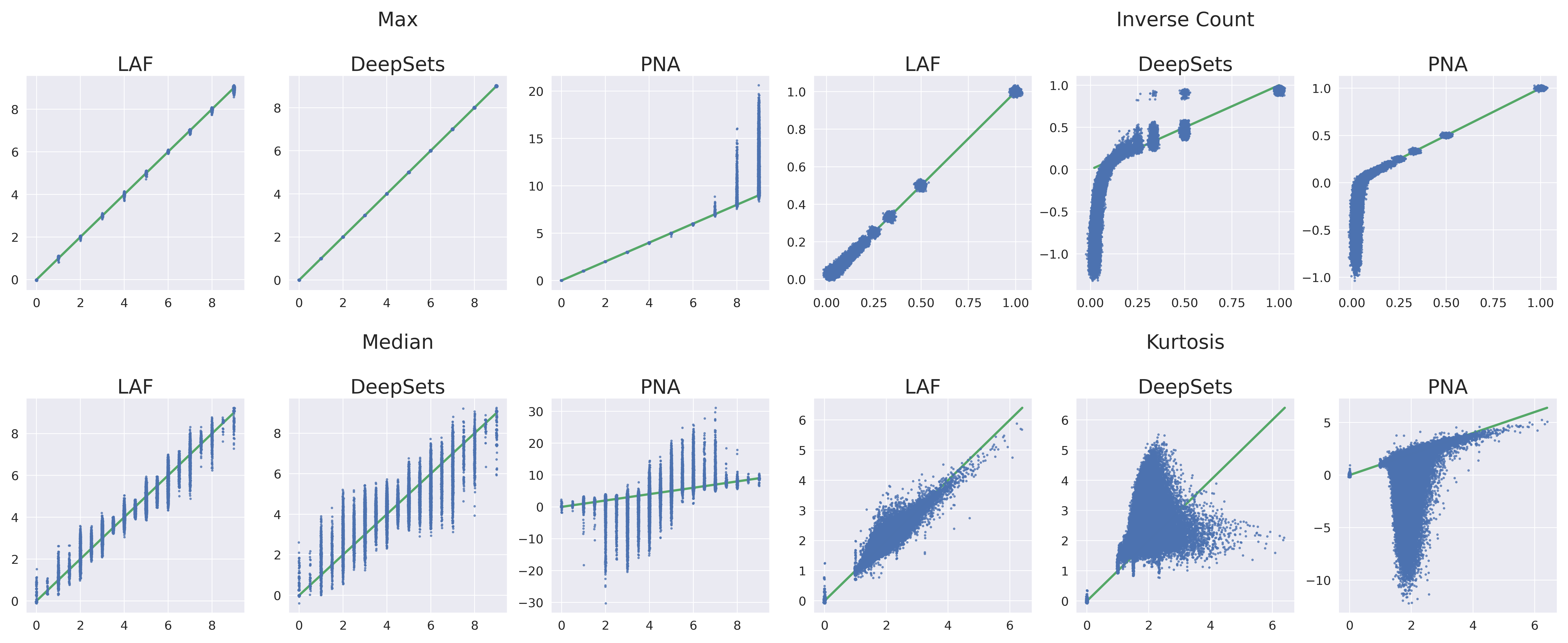}
\figuretag{D.2}
\caption{Scatter plots of the MNIST experiment comparing true (x axis) and predicted (y axis) values with 50 as maximum test set size. The target aggregations are \textit{max} (up-left), \textit{inverse count} (up-right), \textit{median} (bottom-left) and \textit{kurtosis} (bottom-right).}
\label{fig:scatter}
\end{figure*}
Figure~\ref{fig:int_all_big} shows the comparison of LAF, DeepSets, PNA, and LSTM in this setting. 
Results are quite similar to those achieved in the scalar setting, indicating that LAF is capable of effectively backpropagating information so as to drive the learning of an appropriate latent representation, while DeepSets, PNA, and LSTM suffer from the same problems seen in aggregating scalars.

Furthermore, Figure~\ref{fig:scatter} provides a qualitative evaluation of the predictions of the LAF, DeepSets, and PNA methods on a representative subset of the target aggregators. The images illustrate the correlation between the true labels and the predictions. LAF predictions are distributed over the diagonal line, with no clear bias. On the other hand, DeepSets and PNA perform generally worse than LAF, exhibiting higher variances. In particular, for inverse count and kurtosis, DeepSets and PNA predictions are condensed in a specific area, suggesting an overfitting on the training set. 

\section{Details of Sections~SetTransformer with LAF aggregation}
\begin{figure}[t!]
\includegraphics[width=0.49\textwidth]{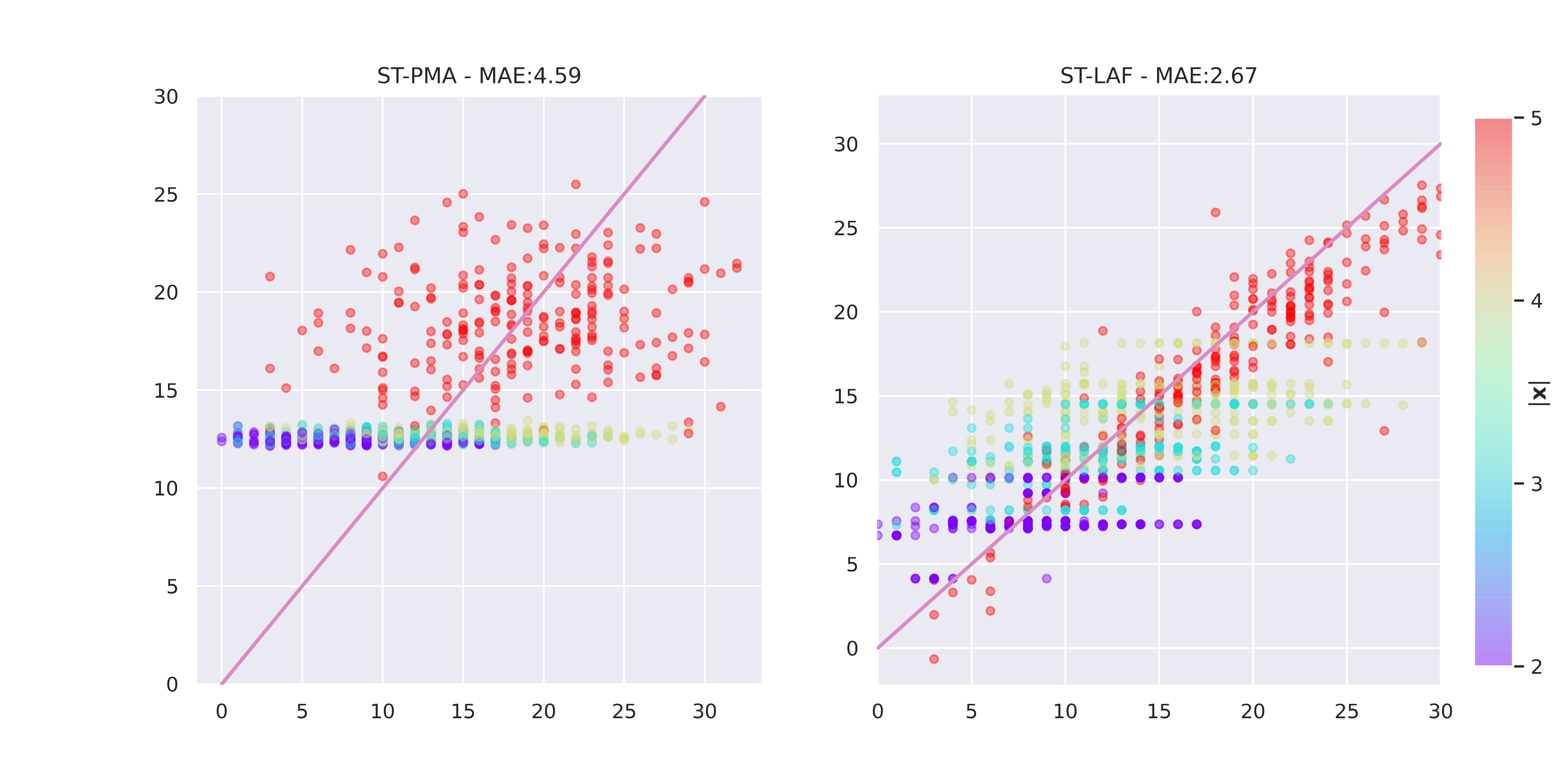}
\figuretag{E.1}
\caption{Distribution of the predicted values for ST-PMA and ST-LAF by set cardinalities. On the x-axis the true labels of the sets, on the y-axis the predicted ones. Different colors represent the sets' cardinalities $|\boldx|$.}
\label{fig:st_pma_laf2}
\end{figure}
We used mini-batches of 64 sets and trained the models for 1,000 epochs. We use Adam as parameter optimizer, setting the initial learning rate to $5e^{-4}$. 
Each element in the dataset is a set of vectors $\boldsymbol{x} = \{x_1,\ldots, x_N \}$, $x_i \in \mathbb{R}^{784}$. 
Network architecture:
\begin{align*}
  \boldsymbol{x}  & \rightarrow  \textsc{Dense(784,300)} \rightarrow \textsc{ReLU}\\
& \rightarrow  \textsc{Dense(300,100)} \rightarrow \textsc{ReLU}\\  
& \rightarrow  \textsc{Dense(100,30)} \rightarrow \textsc{Sigmod}\\  
& \rightarrow  \textsc{SAB(64,4)} \rightarrow \textsc{SAB(64,4)}\\
& \rightarrow  \textsc{PMA$_k$(64,4)} \ \textsc{or} \ \textsc{LAF(10)} \\
& \rightarrow \textsc{Dense(64 $\times$ $k$ or 9, 100)}\rightarrow \textsc{ReLU}\\  
& \rightarrow  \textsc{Dense(100,1)} 
\end{align*}
Please refer to \cite{DBLP:conf/icml/LeeLKKCT19} for
the SAB and PMA details. Figure~\ref{fig:st_pma_laf2} shows the comparison of ST-PMA and ST-LAF for unique sum of MNIST images.
\end{document}